\documentclass[10pt,twocolumn,letterpaper]{article}

\usepackage{iccv}
\usepackage{times}
\usepackage{epsfig}
\usepackage{graphicx}
\usepackage{amsmath}
\usepackage{amssymb}
\usepackage{enumitem}
\usepackage{subcaption}

\usepackage{algorithm}
\usepackage{algorithmic}

\usepackage[pagebackref=true,breaklinks=true,letterpaper=true,colorlinks,bookmarks=false]{hyperref}

\iccvfinalcopy 

\def\iccvPaperID{0000} 

\ificcvfinal\fi
\newcommand{\vect}[1]{\boldsymbol{\mathbf{#1}}}

\makeatletter
\newcommand{\printfnsymbol}[1]{%
	\textsuperscript{\@fnsymbol{#1}}%
}
\makeatother

\begin{document}

\title{RepMLP: Re-parameterizing Convolutions into Fully-connected Layers for Image Recognition}

\author{Xiaohan Ding \textsuperscript{1} \thanks{This work is supported by The National Key Research and Development Program of China (No. 2017YFA0700800), the National Natural Science Foundation of China (No.61925107, No.U1936202) and Beijing Academy of Artificial Intelligence (BAAI). Xiaohan Ding is funded by the Baidu Scholarship Program 2019. This work is done during Xiaohan Ding's internship at MEGVII.} \thanks{Equal contribution.} \quad Chunlong Xia \textsuperscript{2}\printfnsymbol{2} \quad Xiangyu Zhang \textsuperscript{2} \\ Xiaojie Chu \textsuperscript{2} \quad Jungong Han \textsuperscript{3} \quad Guiguang Ding \textsuperscript{1}\thanks{Corresponding author.} \\
	\textsuperscript{1} Beijing National Research Center for Information Science and Technology (BNRist); \\School of Software, Tsinghua University, Beijing, China \\
	\textsuperscript{2} MEGVII Technology \\
	\textsuperscript{3} Computer Science Department, Aberystwyth University, SY23 3FL, UK \\
	\tt\small dxh17@mails.tsinghua.edu.cn \quad xiachunlong2021@outlook.com \quad zhangxiangyu@megvii.com\\
	\tt\small chuxiaojie@megvii.com \quad jungonghan77@gmail.com \quad dinggg@tsinghua.edu.cn\\
}

\maketitle
\ificcvfinal\thispagestyle{empty}\fi

\begin{abstract} 
	We propose RepMLP, a multi-layer-perceptron-style neural network building block for image recognition, which is composed of a series of fully-connected (FC) layers. Compared to convolutional layers, FC layers are more efficient, better at modeling the long-range dependencies and positional patterns, but worse at capturing the local structures, hence usually less favored for image recognition. We propose a structural re-parameterization technique that adds local prior into an FC to make it powerful for image recognition. Specifically, we construct convolutional layers inside a RepMLP during training and merge them into the FC for inference. On CIFAR, a simple pure-MLP model shows performance very close to CNN. By inserting RepMLP in traditional CNN, we improve ResNets by 1.8\% accuracy on ImageNet, 2.9\% for face recognition, and 2.3\% mIoU on Cityscapes with lower FLOPs. Our intriguing findings highlight that combining the global representational capacity and positional perception of FC with the local prior of convolution can improve the performance of neural network with faster speed on both the tasks with translation invariance (\eg, semantic segmentation) and those with aligned images and positional patterns (\eg, face recognition). The code and models are available at \url{https://github.com/DingXiaoH/RepMLP}.
\end{abstract}


\section{Introduction}

The locality of images (\ie, a pixel is more related to its neighbors than the distant pixels) makes Convolutional Neural Network (ConvNet) successful in image recognition, as a conv layer only processes a local neighborhood. In this paper, we refer to this inductive bias as the \textit{local prior}.

On top of that, we also desire the ability to capture the long-range dependencies, which is referred to as the \textit{global capacity} in this paper. Traditional ConvNets model the long-range dependencies by the large receptive fields formed by deep stacks of conv layers \cite{wang2018non}. However, repeating local operations is computationally inefficient and may cause optimization difficulties. Some prior works enhance the global capacity with self-attention-based modules \cite{wang2018non,dosovitskiy2020image,vaswani2017attention}, which has no local prior. For example, ViT \cite{dosovitskiy2020image} is a pure-Transformer model without convolution, which feeds images into the Transformers as a sequence. Due to the lack of local prior as an important inductive bias, ViT needs an enormous amount of training data ($3\times10^8$ images in JFT-300M) to converge.

On the other hand, some images have intrinsic positional prior, which cannot be effectively utilized by a conv layer because it shares parameters among different positions. For example, when someone tries to unlock a cellphone via face recognition, the photo of the face is very likely to be centered and aligned so that the eyes appear at the top and the nose shows at the middle. We refer to the ability to utilize such positional prior as the \textit{positional perception}.

This paper revisits fully-connected (FC) layers to provide traditional ConvNet with global capacity and positional perception. We directly use an FC as the transformation between feature maps to replace conv in some cases. By flattening a feature map, feeding it through FC, and reshaping back, we can enjoy the positional perception (because its parameters are position-related) and global capacity (because every output point is related to every input point). Such an operation is efficient in terms of both the actual speed and theoretical FLOPs, as shown in Table. \ref{table-comparisons}. For the application scenarios where the primary concerns are the accuracy and throughput but not the number of parameters, one may prefer FC-based models to traditional ConvNets. For example, the GPU inference serves usually have tens of GBs of memory, so that the memory occupied by the parameters is minor compared to that consumed by the computations and internal feature maps.
\begin{figure*}
	\begin{center}
		\includegraphics[width=\linewidth]{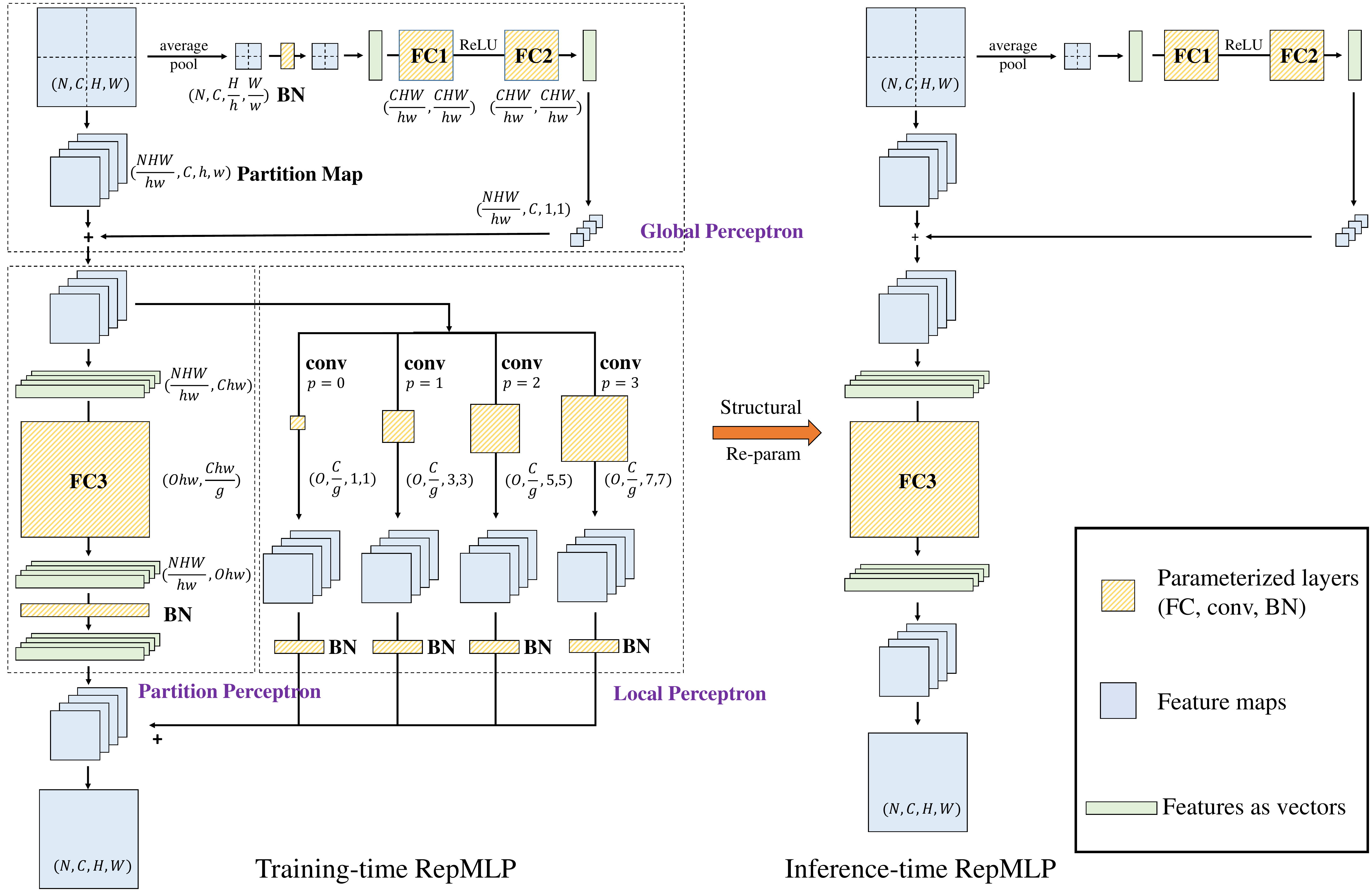}
		\vspace{-0.25in}
		\caption{Sketch of a RepMLP. Here $N,C,H,W$ are the batch size, number of input channels, height and width, $h,w,g,p,O$ are the desired partition height and width, number of groups, padding, and output channels, respectively. The input feature map is split into a set of partitions, and the Global Perceptron adds the correlations among partitions onto each partition. Then the Local Perceptron captures the local patterns with several conv layers, and the Partition Perceptron models the long-range dependencies. This sketch assumes $N=C=1$,$H=W$,$\frac{H}{w}=\frac{W}{w}=2$ (\ie, a channel is split into four partitions) for the better readability. We assume $h,w>7$ so that the Local Perceptron has conv branches of kernel size $1,3,5,7$. The shapes of parameter tensors are shown alongside FC and conv layers. Via structural re-parameterization, the training-time block with conv and BN layers is equivalently converted into a three-FC block, which is saved and used for inference.}
		\label{fig-arch}
		\vspace{-0.25in}
	\end{center}
\end{figure*}

However, an FC has no local prior because the spatial information is lost. In this paper, we propose to incorporate local prior into FC with a \textit{structural re-parameterization} technique. Specifically, we construct conv and batch normalization (BN) \cite{ioffe2015batch} layers parallel to the FC during training, then merge the trained parameters into the FC to reduce the number of parameters and latency for inference. Based on that, we propose a re-parameterized multi-layer perceptron (RepMLP). As shown in Fig. \ref{fig-arch}, the training-time RepMLP has FC, conv, and BN layers but can be equivalently converted into an inference-time block with only three FC layers. The meaning of structural re-parameterization is that the training-time model has a set of parameters while the inference-time model has another set, and we \textit{parameterize the latter with the parameters transformed from the former}. Note that we do not derive the parameters before each inference. Instead, we convert it \textit{once for all}, and then the training-time model can be discarded.

Compared to conv, RepMLP runs faster under the same number of parameters and has global capacity and positional perception. Compared to a self-attention module \cite{wang2018non,dosovitskiy2020image}, it is simpler and can utilize the locality of images. As shown in our experiments (Table. \ref{table-comparisons}, \ref{table-face}, \ref{table-seg}), RepMLP outperforms the traditional ConvNets in a variety of vision tasks, including \textbf{1)} general classification (ImageNet \cite{deng2009imagenet}), \textbf{2)} task with positional prior (face recognition) and \textbf{3)} task with translation invariance (semantic segmentation).

Our contributions are summarized as follows.
\begin{itemize}[noitemsep,nolistsep,topsep=0pt,parsep=0pt,partopsep=0pt]
	\item We propose to utilize the global capacity and positional perception of FC and equip it with local prior for image recognition.
	\item We propose a simple, platform-agnostic and differentiable algorithm to merge the parallel conv and BN into FC for the local prior without any inference-time costs.
	\item We propose RepMLP, an efficient building block, and show its effectiveness on multiple vision tasks.
\end{itemize}




\section{Related Work}

\subsection{Designs for Global Capacity}

Non-local Network \cite{wang2018non} proposed to model the long-range dependencies via the self-attention mechanism. For each query position, the non-local module first computes the pairwise relations between the query position and all positions to form an attention map and then aggregates the features of all the positions by a weighted sum with the weights defined by the attention map. Then the aggregated features are added to the features of each query position.

GCNet \cite{cao2019gcnet} created a simplified network based on a query-independent formulation, which maintains the accuracy of Non-local Network with less computation. The input to a GC block goes through a global attention pooling, feature transform (a $1\times1$ conv), and feature aggregation. 

Compared to these works, RepMLP is simpler as it uses no self-attention and contains only three FC layers. As will be shown in Table. \ref{table-comparisons}, RepMLP improves the performance of ResNet-50 more than Non-local module and GC block.

\subsection{Structural Re-parameterization}

In this paper, structural re-parameterization refers to constructing the conv and BN layers parallel to an FC for training and then merging the parameters into the FC for inference. The following two prior works can also be categorized into structural re-parameterization.

Asymmetric Convolution Block (ACB) \cite{ding2019acnet} is a replacement for regular conv layers, which uses horizontal (\eg, $1\times3$) and vertical ($3\times1$) conv to strengthen the ``skeleton'' of a square ($3\times3$) conv. Reasonable performance improvements are reported on several ConvNet benchmarks.


RepVGG \cite{ding2021repvgg} is a VGG-like architecture, as its body uses only $3\times3$ conv and ReLU for inference. Such an inference-time architecture is converted from a training-time architecture with identity and $1\times1$ branches.

RepMLP is more related to ACB since they are both neural network building blocks, but our contributions are not about making convolutions stronger but \textit{making MLP powerful} for image recognition as a replacement for regular conv. Besides, the training-time convolutions inside RepMLP may be enhanced by ACB, RepVGG block, or other forms of convolution for further improvements.

\section{RepMLP}



A training-time RepMLP is composed of three parts termed as Global Perceptron, Partition Perceptron and Local Perceptron (Fig. \ref{fig-arch}). In this section, we introduce our formulation, describe every component, and show how to convert a training-time RepMLP into three FC layers for inference, where the key is a simple, platform-agnostic and differentiable method for merging a conv into an FC.

\subsection{Formulation}

In this paper, a feature map is denoted by a tensor $\mathrm{M}\in\mathbb{R}^{N\times C\times H\times W}$, where $N$ is the batch size, $C$ is the number of channels, $H$ and $W$ are the height and width, respectively. We use $\mathrm{F}$ and $\mathrm{W}$ for the kernel of conv and FC, respectively. For the simplicity and ease of re-implementation, we use the same data format as PyTorch \cite{paszke2019pytorch} and formulate the transformations in a pseudo-code style. For example, the data flow through a $K\times K$ conv is formulated as
\begin{equation}
	\mathrm{M}^{(\text{out})} = \text{CONV}(\mathrm{M}^{(\text{in})}, \mathrm{F}, p) \,,
\end{equation}
where $\mathrm{M}^{(\text{out})}\in\mathbb{R}^{N\times O\times H^\prime\times W^\prime}$ is the output feature map, $O$ is the number of output channels, $p$ is the number of pixels to pad, $\mathrm{F}\in\mathbb{R}^{O\times C\times K\times K}$ is the conv kernel (we temporarily assume the conv is dense, \ie, the number of groups is~1). From now on, we assume $H^\prime=H, W^\prime=W$ for the simplicity (\ie, the stride is 1 and $p=\lfloor \frac{K}{2} \rfloor$).

For an FC, let $P$ and $Q$ be the input and output dimensions, $\mathrm{V}^{(\text{in})}\in\mathbb{R}^{N\times P}$ and $\mathrm{V}^{(\text{out})}\in\mathbb{R}^{N\times Q}$ be the input and output, respectively, the kernel is $\mathrm{W}\in\mathbb{R}^{Q\times P}$ and the matrix multiplication (MMUL) is formulated as 
\begin{equation}\label{eq-formulation-v}
	\mathrm{V}^{(\text{out})} = \text{MMUL}(\mathrm{V}^{(\text{in})}, \mathrm{W})=\mathrm{V}^{(\text{in})}\cdot\mathrm{W}^\intercal \,.
\end{equation}

We now focus on an FC that takes $\mathrm{M}^{(\text{in})}$ as input and outputs $\mathrm{M}^{(\text{out})}$. We assume the FC does not change the resolution, \ie, $H^\prime=H, W^\prime=W$. We use $\text{RS}$ (short for ``reshape'') as the function that only changes the shape specification of tensors but not the order of data in memory, which is \textit{cost-free}. The input is first flattened into $N$ vectors of length $CHW$, which is $\mathrm{V}^{(\text{in})}=\text{RS}(\mathrm{M}^{(\text{in})}, (N,CHW))$, multiplied by the kernel $\mathrm{W}(OHW, CHW)$, then the output $\mathrm{V}^{(\text{out})}(N, OHW)$ is reshaped back into $\mathrm{M}^{(\text{out})}(N,O,H,W)$. For the better readability, we omit the RS if there is no ambiguity, 
\begin{equation}
	\mathrm{M}^{(\text{out})}=\text{MMUL}(\mathrm{M}^{(\text{in})},\mathrm{W}) \,.
\end{equation}
Such an FC cannot take advantage of the locality of images as it computes each output point according to every input point, unaware of the positional information.

\subsection{Components of RepMLP}


We do not use FC in the above-mentioned manner because of not only the lack of local prior but also the huge number of parameters, which is $COH^2W^2$. With the common settings, \eg, $H=W=28,C=O=128$ on ImageNet, this single FC would have 10G parameters, which is clearly unacceptable. To reduce the parameters, we propose Global Perceptron and Partition Perceptron to model the inter- and intra-partition dependencies separately.


\textbf{Global Perceptron} splits up the feature map so that different partitions can share parameters. For example, an $(N,C,14,14)$ input can be split into $(4N,C,7,7)$, and we refer to every $7\times7$ block as a \textit{partition}. We use an efficient implementation for such splitting with a single operation of memory re-arrangement. Let $h$ and $w$ be the desired height and width of every partition (we assume $H,W$ are divisible by $h,w$ respectively, otherwise we can simply pad the input), the input $\mathrm{M}\in\mathbb{R}^{N\times C\times H\times W}$ is first reshaped into $(N, C, \frac{H}{h}, h, \frac{W}{w}, w)$. Note that this operation is cost-free as it does not move data in memory. Then we re-arrange the order of axes as $(N,\frac{H}{h}, \frac{W}{w}, C, h, w)$, which moves the data in memory efficiently. For example, it requires only one function call (\textit{permute}) in PyTorch. Then the $(N,\frac{H}{h}, \frac{W}{w}, C, h, w)$ tensor is reshaped (which is cost-free again) as $(\frac{NHW}{hw}, C, h, w)$ (noted as a \textit{partition map} in Fig. \ref{fig-arch}). In this way, the number of parameters required is reduced from $COH^2W^2$ to $COh^2w^2$.

However, splitting breaks the correlations among different partitions of the same channel. In other words, the model will view the partitions separately, totally unaware that they were positioned side by side. To add correlations onto each partition, Global Perceptron \textbf{1)} uses average pooling to obtain a pixel for each partition, \textbf{2)} feeds it though BN and a two-layer MLP, then \textbf{3)} reshapes and adds it onto the partition map. This addition can be efficiently implemented with automatic broadcasting (\ie, implicitly replicating $(\frac{NHW}{hw},C,1,1)$ into $(\frac{NHW}{hw},C,h,w)$) so that every pixel is related to the other partitions. Then the partition map is fed into Partition Perceptron and Local Perceptron. Note that if $H=h,W=w$, we directly feed the input feature map into Partition Perceptron and Local Perceptron without splitting, hence there will be no Global Perceptron.

\textbf{Partition Perceptron} has an FC and a BN layer, which takes the partition map. The output $(\frac{NHW}{hw}, O, h, w)$ is reshaped, re-arranged and reshaped in the inverse order as before into $(N, O, H, W)$. We further reduce the parameters of FC3 inspired by groupwise conv \cite{chollet2017xception,xie2017aggregated}. With $g$ as the number of groups, we formulate the groupwise conv as
\begin{equation}
\mathrm{M}^{(\text{out})} = \text{gCONV}(\mathrm{M}^{(\text{in})}, \mathrm{F}, g, p) \,, \mathrm{F}\in\mathbb{R}^{O\times \frac{C}{g} \times K\times K} \,.
\end{equation}

Similarly, the kernel of \textit{groupwise FC} is $\mathrm{W}\in\mathbb{R}^{Q\times \frac{P}{g}}$, which has $g\times$ fewer parameters. Though groupwise FC is not directly supported by some computing frameworks like PyTorch, it can be alternatively implemented by a groupwise $1\times1$ conv. The implementation is composed of three steps: \textbf{1)} reshaping $\mathrm{V}^{(\text{in})}$ as a ``feature map'' with spatial size of $1\times1$; \textbf{2)} performing $1\times1$ conv with $g$ groups; \textbf{3)} reshaping the output ``feature map'' into $\mathrm{V}^{(\text{out})}$. We formulate the groupwise matrix multiplication (gMMUL) as
\begin{equation}
\begin{aligned}
&\mathrm{M}^\prime=\text{RS}(\mathrm{V}^{(\text{in})}, (N, P, 1, 1)),\quad \mathrm{F}^\prime=\text{RS}(\mathrm{W}, (Q, \frac{P}{g}, 1, 1) \,, \\
&\text{gMMUL}(\mathrm{V}^{(\text{in})}, \mathrm{W}, g) = \text{RS}(\text{gCONV}(\mathrm{M}^\prime, \mathrm{F}^\prime, g, 0), (N, Q)) \,.
\end{aligned}
\end{equation}

\textbf{Local Perceptron} feeds the partition map through several conv layers. A BN follows every conv, as inspired by \cite{ding2019acnet,ding2021repvgg}. Fig. \ref{fig-arch} shows an example of $h,w>7$ and $K=1,3,5,7$. Theoretically, the only constraint on the kernel size $K$ is $K\leq h,w$ (because it does not make sense to use kernels larger than the resolution), but we only use odd kernel sizes as a common practice in ConvNet. We use $K\times K$ just for the simplicity of notation and a non-square conv (\eg, $1\times3$ or $3\times5$) also works. The padding of conv should be configured to maintain the resolution (\eg, $p=0,1,2,3$ for $K=1,3,5,7$, respectively), and the number of groups $g$ should be the same as the Partition Perceptron. The outputs of all the conv branches and Partition Perceptron are added up as the final output.

\subsection{A Simple, Platform-agnostic, Differentiable Algorithm for Merging Conv into FC}
Before converting a RepMLP into three FC layers, we first show how to merge a conv into FC. With the FC kernel $\mathrm{W}^{(1)}(Ohw,Chw)$, conv kernel $\mathrm{F}(O,C,K,K)$ ($K\leq h,w$) and padding $p$, we desire to construct $\mathrm{W}^\prime$ so that
\begin{equation}
\begin{aligned}
&\text{MMUL}(\mathrm{M}^{(\text{in})},\mathrm{W}^\prime) \\
&=\text{MMUL}(\mathrm{M}^{(\text{in})},\mathrm{W}^{(1)}) + \text{CONV}(\mathrm{M}^{(\text{in})},\mathrm{F},p) \,.
\end{aligned}
\end{equation}

We note that for any kernel $\mathrm{W}^{(2)}$ of the same shape as $\mathrm{W}^{(1)}$, the additivity of MMUL ensures that
\begin{equation}
\begin{aligned}
&\text{MMUL}(\mathrm{M}^{(\text{in})}, \mathrm{W}^{(1)}) + \text{MMUL}(\mathrm{M}^{(\text{in})}, \mathrm{W}^{(2)}) \\
&= \text{MMUL}(\mathrm{M}^{(\text{in})}, \mathrm{W}^{(1)} + \mathrm{W}^{(2)}) \,,
\end{aligned}
\end{equation}
so we can merge $\mathrm{F}$ into $\mathrm{W}^{(1)}$ as long as we manage to construct $\mathrm{W}^{(\mathrm{F},p)}$ of the same shape as $\mathrm{W}^{(1)}$ which satisfies
\begin{equation}
	\text{MMUL}(\mathrm{M}^{(\text{in})}, \mathrm{W}^{(\mathrm{F},p)}) = \text{CONV}(\mathrm{M}^{(\text{in})},\mathrm{F},p) \,.
\end{equation}
Obviously, $\mathrm{W}^{(\mathrm{F},p)}$ must exist, since a conv can be viewed as a sparse FC that shares parameters among spatial positions, which is exactly the source of its translation invariance, but it is not obvious to construct it with given $\mathrm{F}$ and $p$. As modern computing platforms use different algorithms of convolution (\eg, im2col-\cite{im2col}, Winograd- \cite{winograd}, FFT-\cite{fft-conv}, MEC-\cite{cho2017mec}, and sliding-window-based) and the memory allocation of data and implementations of padding may be different, a means for constructing the matrix on a specific platform may not work on another platform. In this paper, we propose a simple and \textit{platform-agnostic} solution.

As discussed above, for \textit{any} input $\mathrm{M}^{(\text{in})}$ and conv kernel $\mathrm{F}$, padding $p$, there exists an FC kernel $\mathrm{W}^{(\mathrm{F},p)}$ such that
\begin{equation}
	\mathrm{M}^{(\text{out})} = \text{CONV}(\mathrm{M}^{(\text{in})}, \mathrm{F}, p) = \text{MMUL}(\mathrm{M}^{(\text{in})}, \mathrm{W}^{(\mathrm{F},p)}) \,.
\end{equation}
With the formulation used before (Eq. \ref{eq-formulation-v}), we have
\begin{equation}\label{eq-middle}
	\mathrm{V}^{(\text{out})} = \mathrm{V}^{(\text{in})}\cdot\mathrm{W}^{(\mathrm{F},p)\intercal} \,.
\end{equation}
We insert an identity matrix $\mathrm{I}$ $(Chw, Chw)$ and use the associative law
\begin{equation}
	\mathrm{V}^{(\text{out})} = \mathrm{V}^{(\text{in})}\cdot (\mathrm{I} \cdot \mathrm{W}^{(\mathrm{F},p)\intercal}) \,.
\end{equation}

We note that because $\mathrm{W}^{(\mathrm{F},p)}$ is constructed with $\mathrm{F}$, $\mathrm{I} \cdot \mathrm{W}^{(\mathrm{F},p)\intercal}$ is a convolution with $\mathrm{F}$ on a feature map $\mathrm{M}^{(\mathrm{I})}$ which is reshaped from $\mathrm{I}$. With explicit RS, we have
\begin{equation}
	\mathrm{M}^{(\mathrm{I})} = \text{RS}(\mathrm{I}, (Chw, C, h, w)) \,,
\end{equation}
\begin{equation}\label{eq-second-last}
	\mathrm{I}\cdot\mathrm{W}^{(\mathrm{F},p)\intercal} = \text{CONV}(\mathrm{M}^{(\mathrm{I})}, \mathrm{F}, p) \,,
\end{equation}
\begin{equation}\label{eq-last}
	\mathrm{V}^{(\text{out})} = \mathrm{V}^{(\text{in})} \cdot \text{RS}(\mathrm{I}\cdot\mathrm{W}^{(\mathrm{F},p)\intercal}, (Chw, Ohw)) \,.
\end{equation}
Comparing Eq. \ref{eq-middle} with Eq. \ref{eq-second-last}, \ref{eq-last}, we have
\begin{equation}\label{eq-final}
	\mathrm{W}^{(\mathrm{F},p)} = \text{RS}(\text{CONV}(\mathrm{M}^{(\mathrm{I})}, \mathrm{F}, p), (Chw, Ohw))^\intercal \,.
\end{equation}
Which is exactly the expression we desire for constructing $\mathrm{W}^{(\mathrm{F},p)}$ with $\mathrm{F}, p$. In short, the equivalently FC kernel of a conv kernel is the result of convolution on an identity matrix with proper reshaping. Better still, the conversion is efficient and differentiable, so one may derive the FC kernel during training and use it in the objective function (\eg, for penalty-based pruning \cite{han2015learning,molchanov2016pruning}). The expression and code for the groupwise case are derived in a similar way and provided in the supplementary material.

\subsection{Converting RepMLP into Three FC Layers}

To use the theory presented above, we need to first eliminate the BN layers by equivalently fusing them into the preceding conv layers and FC3. Let $\mathrm{F}\in\mathbb{R}^{O\times \frac{C}{g}\times K\times K}$ be the conv kernel, $\vect{\mu},\vect{\sigma},\vect{\gamma},\vect{\beta}\in\mathbb{R}^{O}$ be the accumulated mean, standard deviation and learned scaling factor and bias of the following BN, we construct the kernel $\mathrm{F}^\prime$ and bias $\mathbf{b}^\prime$ as
\begin{equation}\label{eq-fuse-bn}
\mathrm{F}^\prime_{i,:,:,:} = \frac{\vect{\gamma}_i}{\vect{\sigma}_i}\mathrm{F}_{i,:,:,:} \,,\quad \mathbf{b}^\prime_i = -\frac{\vect{\mu}_i \vect{\gamma}_i}{\vect{\sigma}_i} + \vect{\beta}_i \,.
\end{equation}
Then it is easy to verify the equivalence:
\begin{equation}
\begin{aligned}
&\frac{\vect{\gamma}_i}{\vect{\sigma}_i}(\text{CONV}(\mathrm{M}, \mathrm{F}, p)_{:,i,:,:} - \vect{\mu}_i) + \vect{\beta}_i \\
&= \text{CONV}(\mathrm{M}, \mathrm{F}^\prime, p)_{:,i,:,:} + \mathbf{b}^\prime_i \,, \forall 1\leq i \leq O \,,
\end{aligned}
\end{equation}
where the left side is the original computation flow of a conv-BN, and the right is the constructed conv with bias.
 
The 1D BN and FC3 of Partition Perceptron are fused in a similar way into $\hat{\mathrm{W}}\in\mathbb{R}^{Ohw\times\frac{Chw}{g}}$, $\hat{\mathbf{b}}\in\mathbb{R}^{Ohw}$. Then we convert every conv via Eq. \ref{eq-final} and add the resultant matrix onto $\hat{\mathrm{W}}$. The biases of conv are simply replicated by $hw$ times (because all the points on the same channel share a bias value) and added onto $\hat{\mathbf{b}}$. Finally, we obtain a single FC kernel and a single bias vector, which will be used to parameterize the inference-time FC3.


The BN in Global Perceptron is also removed because the removal is equivalent to applying an affine transformation before FC1, which can be absorbed by FC1 as two sequential MMULs can be merged into one. The formulas and code are provided in the supplementary material.

\subsection{RepMLP-ResNet}

\begin{figure}
	\begin{center}
		\includegraphics[width=\linewidth]{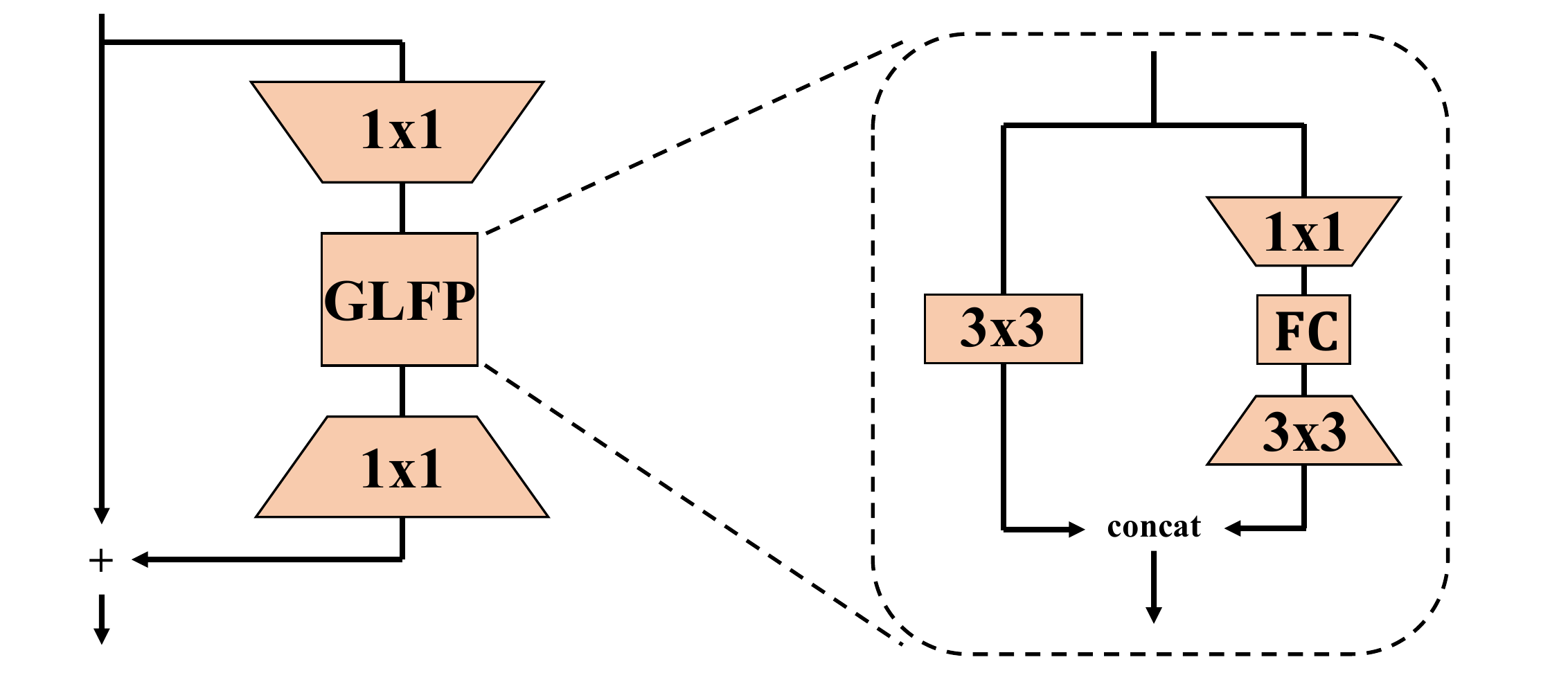}
		\vspace{-0.20in}
		\caption{Illustration of GLFP Module (courtesy of \cite{xiapatent}), which is relevant to our RepMLP Bottleneck designed for ResNets (Fig. \ref{fig-repmlp-resnet}).}
		\label{fig-GLFP}
		\vspace{-0.3in}
	\end{center}
\end{figure}

The design of RepMLP and the methodology of re-parameterizing conv into FC are generic hence may be used in numerous models including traditional CNNs and the concurrently proposed all-MLP models, \eg, MLP-Mixer \cite{tolstikhin2021mlp}, ResMLP \cite{touvron2021resmlp}, gMLP \cite{liu2021pay}, AS-MLP \cite{lian2021mlp}, \etc. In this paper, we use RepMLP in ResNet for most of our experiments because this work was finished before the publicity of all the above-mentioned all-MLP models. The application of RepMLP on the all-MLP models is scheduled as our future work.

\begin{figure}
	\begin{center}
		\includegraphics[width=\linewidth]{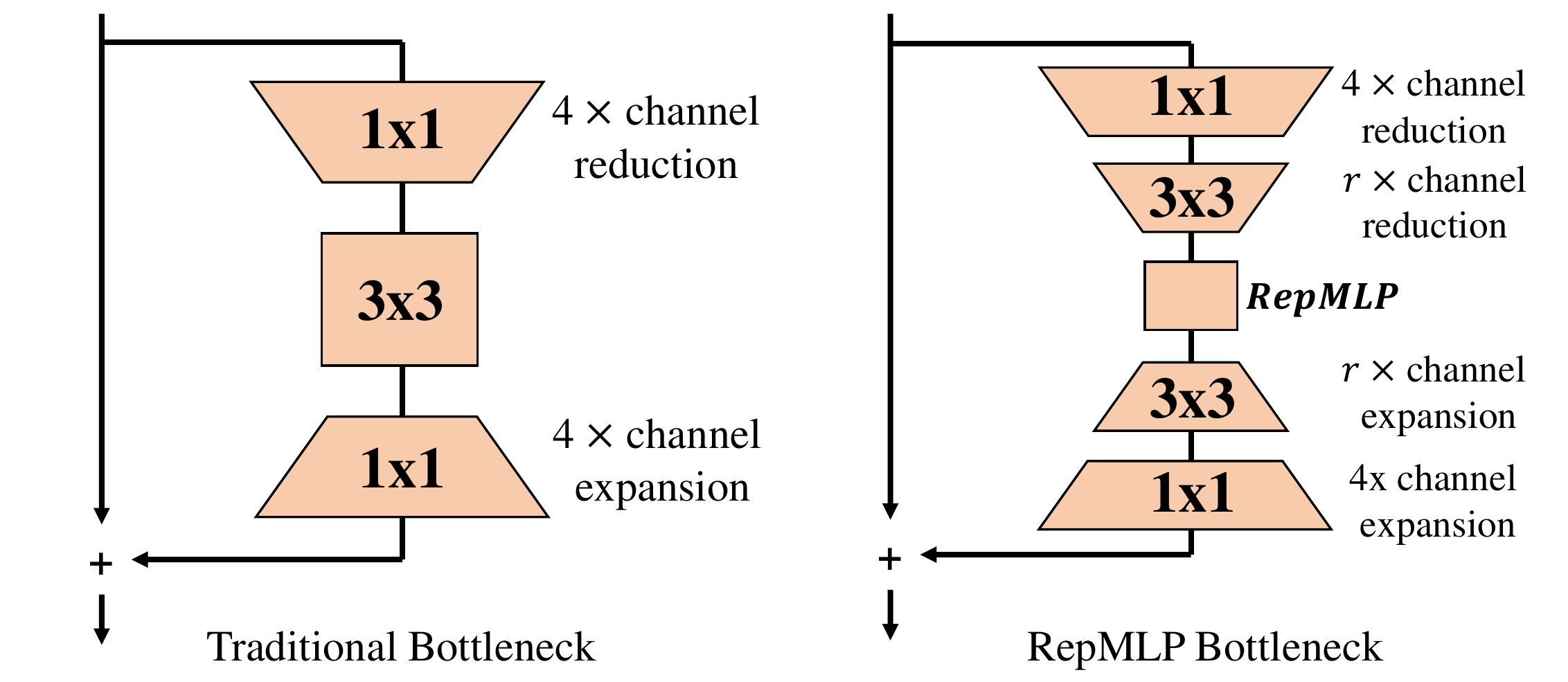}
		\vspace{-0.20in}
		\caption{Sketch of a RepMLP Bottleneck.}
		\label{fig-repmlp-resnet}
		\vspace{-0.3in}
	\end{center}
\end{figure}

In order to use RepMLP in ResNet, we follow the bottleneck \cite{he2016deep} design principle of ResNet-50 to reduce the channels by $4\times$ via $1\times1$ conv. Moreover, we further perform $r\times$ channel reduction before RepMLP and $r\times$ channel expansion afterwards via $3\times3$ conv. The whole block is termed as RepMLP Bottleneck (Fig. \ref{fig-repmlp-resnet}). For a specific stage, we replace all the stride-1 bottlenecks with RepMLP Bottlenecks and keep the original stride-2 (\ie, the first) bottleneck.

The design of RepMLP Bottleneck is relevant to GLFP Module \cite{xiapatent}, which uses a bottleneck structure with $1\times1$, $3\times3$ conv and FC for human face recognition, but the differences are significant. \textbf{1)} GLFP directly flattens the input feature maps as vectors then feeds them into the FC layer, which is novel and insightful but may be inefficient on tasks with large input resolution such as ImageNet classification and semantic segmentation. In contrast, RepMLP partitions the input feature maps and use Global Perceptron to add the global information. \textbf{2)} GLFP uses a $3\times3$ conv branch parallel to the $1\times1$-FC-$3\times3$ branch to capture the local patterns. Unlike the Local Perceptron of RepMLP that can be merged into the FC for inference, the conv branch of GLFP is essential for both training and inference. \textbf{3)} Some differences in the topology (e.g., addition v.s. concatenation). It should be noted again that the core contribution of this paper is not the solution to insert RepMLP into ResNet but the methodology of re-parameterizing conv into FC and the three components of RepMLP.





\section{Experiments}

\subsection{Pure MLP and Ablation Studies}
\begin{figure}
	\begin{center}
		\includegraphics[width=\linewidth]{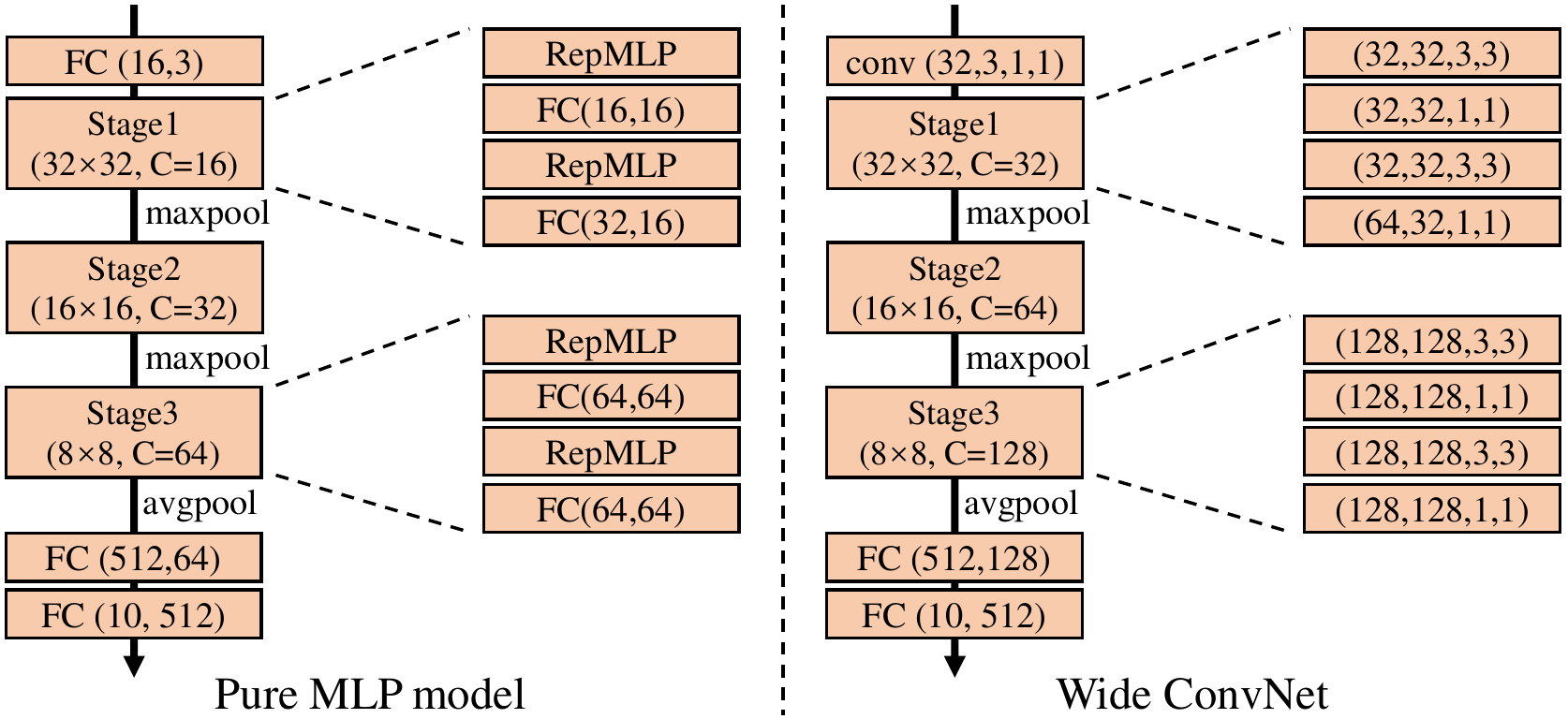}
		\vspace{-0.25in}
		\caption{The pure MLP model and the convolutional counterpart. The stage1 and stage3 are displayed in detail. Taking stage1 for example, $32\times32$ is the resolution, $C=16$ is the number of output channels (except the last layer). Left: FC(32,16) is the kernel size, suggesting that this FC (equivalent to a $1\times1$ conv) projects 16 channels into 32 channels; all the RepMLPs are configured with $g=2$, $h=w=8$. Right: the convolutional counterpart uses $3\times3$ conv. A BN follows every conv and a ReLU follows every RepMLP or conv-BN sequence.}
		\label{fig-puremlp}
		\vspace{-0.25in}
	\end{center}
\end{figure}

We first verify the effectiveness of RepMLP by testing a pure MLP model on CIFAR-10. More precisely, since an FC is equivalent to a $1\times1$ conv, by ``pure MLP'' we means no usage of conv kernels bigger than $1\times1$. We interleave RepMLP and regular FC ($1\times1$ conv) to construct three stages and downsample by max pooling, as shown in Fig.~\ref{fig-puremlp}, and construct a ConvNet counterpart for comparison by replacing the RepMLPs with $3\times3$ conv. For the comparable FLOPs, the channels of the three stages are 16,32,64 for the pure MLP and 32,64,128 for the ConvNet, so the latter is named Wide ConvNet. We adopt the standard data augmentation \cite{he2016deep}: padding to $40\times40$, random cropping and left-right flipping. The models are trained with a batch size of 128 and a cosine learning rate annealing from 0.2 to 0 in 100 epochs. As shown in Table. \ref{table-puremlp}, the pure MLP model reaches 91.11\% accuracy with only 52.8M FLOPs. Not surprisingly, the pure MLP model does not outperform the Wide ConvNet, motivating us to combine RepMLP and traditional ConvNet. 

Then we conduct a series of ablation studies. \textbf{A)} We also report the FLOPs of the MLP before the conversion, which still contains conv and BN layers. The FLOPs is much higher though the extra parameters are marginal, which shows the significance of structural re-parameterization. \textbf{B)} ``w/o Local'' is a variant with no Local Perceptron, and the accuracy is 8.5\% lower, which shows the significance of local prior. \textbf{C)} ``w/o Global'' removes FC1 and FC2 and directly feed the partition map into Local Perceptron and Partition Perceptron. \textbf{D)} ``FC3 as conv9'' replaces FC3 with a conv ($K=9$ and $p=4$, so that its receptive field is larger than FC3) followed by BN to compare the representational capacity of FC3 to a regular conv. Though the comparison is biased towards conv because its receptive field is larger, its accuracy is 3.5\% lower, which validates that FC is more powerful than conv since a conv is a degraded FC. \textbf{E)} ``RepMLP as conv9'' directly replaces the RepMLP with a $9\times9$ conv and BN. Compared to D, its accuracy is lower as it has no Global Perceptrons. 

\setlength{\tabcolsep}{4pt}
\begin{table}
	\caption{Top-1 accuracy, FLOPs and parameters of pure MLP and ConvNet on CIFAR-10. }
	\label{table-puremlp}
	\vspace{-0.2in}
	\begin{center}
		\small
		\begin{tabular}{lcccccccc}
			\hline
			Model					&	Acc		&	 FLOPs (M)	&	Params (M) 	\\
			\hline
			Pure MLP				&	91.11	&	52.8		&	22.41	\\
			A) before conversion	&	91.11	&	118.9		&	22.91	\\	
			B) w/o Local			&	82.64	&	52.8		&	22.41	\\	
			C) w/o Global			&	89.61	&	52.5		&	22.08	\\
			D) FC3 as conv9			&	87.64	&	66.2		&	0.81		\\
			E) RepMLP as conv9		&	87.29	&	65.8		&	0.48		\\
			\hline			
			Wide ConvNet	&	91.99	&	65.1		&	0.50		\\
			\hline
		\end{tabular}
	\end{center}
	\vspace{-0.2in}
\end{table}
\setlength{\tabcolsep}{1.4pt}


\subsection{RepMLP-ResNet for ImageNet Classification}


We take ResNet-50 \cite{he2016deep} (the torchvision version \cite{torch-model}) as the base architecture to evaluate RepMLP as a building block in traditional ConvNet. For the fair comparison, all the models are trained with identical settings in 100 epochs: global batch size of 256 on 8 GPUs, weight decay of $10^{-4}$, momentum of 0.9, and cosine learning rate annealing from 0.1 to 0. We use mixup \cite{zhang2017mixup} and a data augmentation pipeline of Autoaugment \cite{cubuk2019autoaugment}, random cropping and flipping. All the models are evaluated with single central crop and the speed is tested on the same 1080Ti GPU with a batch size of 128 and measured in examples/second. For the fair comparison, the RepMLPs are converted and all the original conv-BN structures of every model are also converted into conv layers with bias for the speed tests. 

As a common practice, we refer to the four residual stages of ResNet-50 as c2, c3, c4, c5, respectively. With $224\times224$ input, the output resolutions of the four stages are $56, 28, 14, 7$, and the $3\times3$ conv layers in the four stages have $C=O=64,128,256,512$, respectively. To replace the big $3\times3$ conv layers with RepMLP, we use $h=w=7$ and three conv branches in the Local Perceptron with $K=1,3,5$. 

We begin by using RepMLP in c4 only and varying the hyper-parameters $r$ and $g$ to test how they influence the accuracy, speed, and number of parameters (Table. \ref{table-c4-gr}). Notably, with violent 8$\times$ reduction (so that the input and output channels of RepMLP is $256/8=32$), RepMLP-Res50 has fewer parameters and run 10\% faster than ResNet-50. The comparison between the first two rows suggest that the current groupwise $1\times1$ conv is inefficient, as the parameters increase by 59\% but the speed decreases by only 0.7\%. Further optimizations on groupwise $1\times1$ conv may make RepMLP more efficient. In the following experiments, we use $r=2$ or 4 and $g=4$ or 8 for the better trade-off.

\setlength{\tabcolsep}{4pt}
\begin{table}
	\caption{Results with $224\times224$ input and different $r,g$ in c4 only. The speed is in examples/second.}
	\label{table-c4-gr}
	\vspace{-0.25in}
	\begin{center}
		\small
		\begin{tabular}{lccccccc}
			\hline
			&	$r$			&	 $g$ 		&	Top-1 acc	&	Speed	&	Params (M)	\\
			\hline
			RepMLP-Res50	&4				&	8					&	78.13		&	671		&	30.87	\\
			RepMLP-Res50	&4				&	2					&	78.22		&	666		&	49.31	\\
			RepMLP-Res50	&8				&	8					&	77.79		&	759		&	25.02	\\
			RepMLP-Res50	&2				&	8					&	78.60		&	639		&	52.77	\\
			Original Res50	&-				&	-					&	77.19		&	689		&	25.53	\\
			\hline
		\end{tabular}
	\end{center}
	\vspace{-0.15in}
\end{table}
\setlength{\tabcolsep}{1.4pt}

\setlength{\tabcolsep}{4pt}
\begin{table}
	\caption{Using RepMLP in different stages of ResNet-50 with $224\times224$ input. The speed is in examples/second.}
	\label{table-stages}
	\vspace{-0.25in}
	\begin{center}
		\small
		\begin{tabular}{lccccccc}
			\hline
			&	\text{c2}		&	 \text{c3} 		&	\text{c4}	&	\text{c5}	&	Top-1 acc	&	Speed	&	Params (M)	\\
			\hline
			RepMLP-Res50&\checkmark		&	\checkmark		&	\checkmark	&	\checkmark	&	78.32		&	574		&	74.46	\\
			RepMLP-Res50&\checkmark		&	\checkmark		&	\checkmark	&				&	78.60		&	575		&	66.97	\\
			RepMLP-Res50&				&	\checkmark		&	\checkmark	&	\checkmark	&	78.30		&	632		&	48.35	\\
			RepMLP-Res50&				&	\checkmark		&	\checkmark	&				&	78.55		&	636		&	40.87	\\
			RepMLP-Res50&				&	\checkmark		&				&				&	78.09		&	644		&	35.52	\\
			RepMLP-Res50&				&					&	\checkmark	&				&	78.13		&	671		&	30.87	\\
			Original Res50&				&					&				&				&	77.19		&	689		&	25.53	\\
			\hline
		\end{tabular}
	\end{center}
	\vspace{-0.15in}
\end{table}
\setlength{\tabcolsep}{1.4pt}

We continue to test RepMLP in different stages. Specifically, we set $g=8$ and $r=2,2,4,4$ for c2,c3,c4,c5, respectively, for the reasonable model sizes. Table. \ref{table-stages} shows that replacing the original bottlenecks with RepMLP Bottlenecks causes very minor slowdown, and the accuracy is significantly improved. Using RepMLP only on c4 brings only 5M more parameters but 0.94\% higher accuracy, and using RepMLP in c3 and c4 offers the best trade-off. It also suggests that RepMLP should be combined with traditional conv for the best performance, as using it in all the four stages delivers lower accuracy than c2+c3+c4 and c3+c4. We use RepMLP in c3+c4 in the following experiments.

The comparisons to the larger traditional ConvNets with higher input resolution (Table. \ref{table-comparisons}) further justifies the effectiveness of RepMLP and offers some interesting discoveries. When trained and tested with $320\times320$ inputs, we use RepMLP with $h=w=10$ and the Local Perceptron has four branches with $K=1,3,5,7$. We also vary the number of groups to generate three models with different sizes. For example, g8/16 means that $g=8$ for c3 and 16 for c4. As two classic models for modeling the long-range dependencies, we construct the Non-local \cite{wang2018non} and GC \cite{cao2019gcnet} counterparts following the instructions in the original papers, and the models are trained with the identical settings. We also present the well-known EfficientNet \cite{efficientnet} series as a strong baseline trained with the identical settings again. We have the following observations.

\textbf{1)} Compared to the traditional ConvNets with comparable numbers of parameters, the FLOPs of RepMLP-Res50 is much lower and the speed is faster. For example, compared to ResNet-101 with $224\times224$ inputs, RepMLP-Res50 has only 50\% FLOPs and 4M fewer parameters, runs 50\% faster, but their accuracies are the same. With $320\times320$ inputs, RepMLP-Res50 outperforms in accuracy, speed, and FLOPs by a large margin. Additionally, the improvements of ResNet-50 should not be simply attributed to the increased depth because it is still shallower than ResNet-101. \textbf{2)} Increasing the parameters of RepMLPs causes very minor slowdown. From RepMLP-Res50-g8/16 to RepMLP-Res50-g4/8, the parameters increase by 47\%, but the FLOPs increases by only 3.6\% and the speed is lowered by only 2.2\%. This property is particularly useful for high-throughput inference on large-scale servers, where the throughput and accuracy are our major concerns while the model size is not. \textbf{3)} Compared to Nonlocal and GC, the speed of RepMLP-Res50 is almost the same, but the accuracy is around 1\% higher. \textbf{4)} Compared to EfficientNets, which are actually not efficient on GPU, RepMLP-Res50 outperforms in both the speed and accuracy.


\setlength{\tabcolsep}{4pt}
\begin{table*}
	\caption{Comparisons with traditional ConvNets on ImageNet all trained with the identical settings. The speed is tested on the same 1080Ti with a batch size of 128. The input resolutions of the EfficientNets are different because they are fixed as the structural hyper-parameters.}
	\label{table-comparisons}
	\vspace{-0.2in}
	\begin{center}
		\small
		\begin{tabular}{lcccccccc}
			\hline
			Model			&	Input resolution&	Top-1 acc&	Speed (examples/s)	&	FLOPs (M)	&	Params (M)	\\
			\hline
			ResNet-50		&	224				&	77.19		&	689		&	4089	&	25.53			\\
			ResNet-101		&	224				&	78.55		&	421		&	7801	&	44.49			\\
			RepMLP-Res50	&	224				&	78.55		&	636		&	3890	&	40.87			\\	
			\hline
			ResNet-50		&	320				&	78.03		&	344		&	8343	&	25.53			\\
			ResNet-101		&	320				&	79.40		&	213		&	15919	&	44.49			\\
			
			RepMLP-Res50-g8/16&	320				&	79.76		&	312		&	8057	&	59.22			\\		
			RepMLP-Res50-g8/8&	320				&	79.84		&	311		&	8108	&	72.02			\\		
			RepMLP-Res50-g4/8&	320				&	80.07		&	305		&	8354	&	87.38			\\		

			NL-Res50	&	320				&	78.95		&	316		&	9182	&	27.63			\\
			GC-Res50	&	320				&	78.93		&	312		&	8351	&	28.05			\\
			\hline
			EfficientNet-B1	&	240				&	75.76		&	512		&	686		&	7.76	\\
			EfficientNet-B2	&	260				&	76.46		&	396		&	993		&	9.07	\\
			EfficientNet-B3	&	300				&	78.17		&	228		&	1827	&	12.18	\\
			\hline
		\end{tabular}
	\end{center}
	\vspace{-0.25in}
\end{table*}
\setlength{\tabcolsep}{1.4pt}

\begin{figure}
	\begin{subfigure}{0.47\linewidth}
		\includegraphics[width=\linewidth,page=2]{fc_converted_6_6.pdf} 
	\end{subfigure}
	\begin{subfigure}{0.47\linewidth}
		\includegraphics[width=\linewidth,page=1]{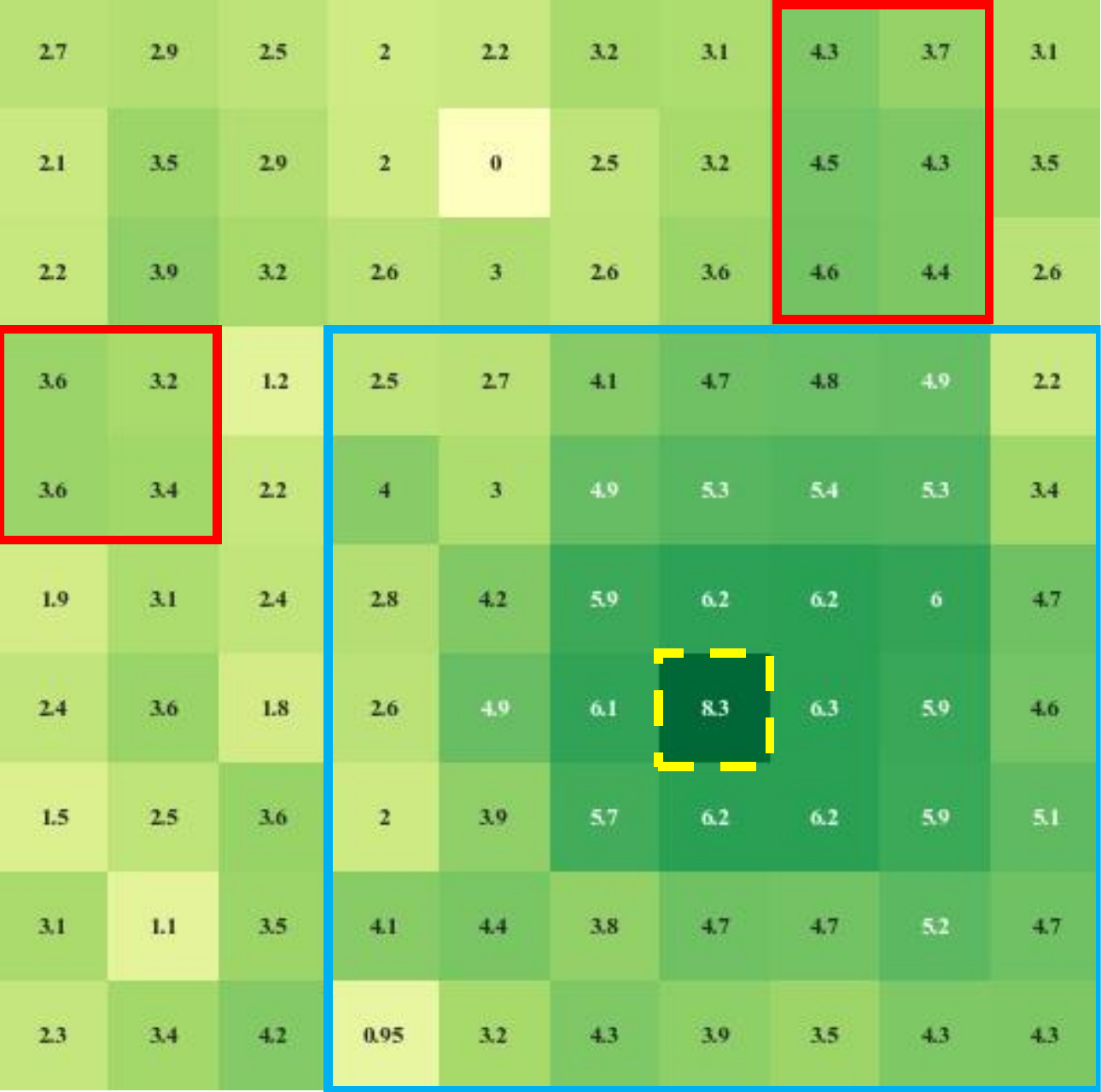} 
	\end{subfigure}
	\vspace{-0.05in}
	\caption{FC weights sampled from the FC3 of the first RepMLP in c3 of RepMLP-Res50-g8/8. The left is the original training-time FC3 and the right is the inference-time FC3 merged with Local Perceptron. Specifically, we reshape the kernel of FC3 into $\bar{\mathrm{W}}(O,h,w,\frac{C}{g},h,w)$, which is $(64, 10, 10, 8, 10, 10)$, then sample the weights related to the first input channel and the (6,6) point (marked by a dashed square) on the first output channel, which is $\bar{\mathrm{W}}_{0,6,6,0,:,:}$. Then we take the absolute value, normalize by the minimum of the whole matrix, and take the natural logarithm for the better readability. A point with darker color indicates the FC considers the corresponding position on the input channel more related to the output point at (6,6).}
	\label{fig-visualized}
	\vskip -0.1in
\end{figure}

We visualize the weights of FC3 in Fig. \ref{fig-visualized}, where the sampled output point (6,6) is marked by a dashed square. The original FC3 has no local prior as the marked point and the neighborhood have no larger values than the others. But after merging the Local Perceptron, the resultant FC3 kernel has larger values around the marked point, suggests that the model focuses more on the neighborhood, which is expected. Besides, the global capacity is not lost because some points (marked by red rectangles) outside the largest conv kernel ($7\times7$ in this case, marked by a blue square) still have larger values than the points inside.

We also present another design of bottleneck (RepMLP Light Block) in the Appendix, which uses no $3\times3$ conv but only $1\times1$ for 8$\times$ channel reduction/expansion. Compared to the original ResNet-50, it achieves comparable accuracy (77.14\% \vs 77.19\%) with 30\% lower FLOPs and 55\% faster speed.

%
%
%

\subsection{Face Recognition}

Unlike conv, FC is not translation-invariant, making RepMLP particularly effective for images with positional prior, \ie, human faces. The dataset we use for training is MS1M-V2, a large-scale face dataset with 5.8M images from 85k celebrities. It is a semi-automatic refined version of the MS-Celeb-1M dataset \cite{guo2016ms} which consists of 1M photos from 100k identities and has many noisy images and wrong ID labels. We use MegaFace \cite{kemelmacher2016megaface} for evaluation, which includes 1M images of 60k identities as the gallery set and 100k images of 530 identities from FaceScrub as the probe set. It is also a refined version by manual clearing. We use $96\times96$ inputs for both training and evaluation.

Apart from MobileFaceNet \cite{chen2018mobilefacenets} as a well-known baseline, which was originally designed for low-power devices, we also use a customized ResNet (referred to as FaceResNet in this paper) as a stronger baseline. Compared to a regular ResNet-50, the numbers of blocks in c2,c3,c4,c5 are reduced from 3,4,6,3 to 3,2,2,2, the widths are reduced from 256,512,1024,2048 to 128,256,512,1024, and the channels of $3\times3$ are increased from 64,128,256,512 to 128,256,512,1024. In other words, the $1\times1$ conv layers in residual blocks do not reduce or expand the channels. Because the input resolution is $96\times96$, the spatial sizes of c2,c3,c4,c5 are 24,12,6,3, respectively. For the RepMLP counterpart, we modify FaceResNet by replacing the stride-1 bottlenecks of c2,c3,c4 (\ie, the last two bottlenecks of c2 and the last blocks of c3,c4) by RepMLP Bottlenecks with $h=w=6,r=2,g=4$.

For training, we use a batch size of 512, momentum of 0.9, AM-Softmax loss \cite{wang2018additive}, and weight decay following \cite{chen2018mobilefacenets}. All the models are trained for 420k iterations with a learning rate beginning with 0.1 and divided by 10 at 252k, 364k and 406k iterations. For evaluation, we report the top-1 accuracy on MegaFace. Table. \ref{table-face} shows that FaceResNet delivers higher accuracy than MobileFaceNet but runs slower, while RepMLP-FaceRes outperforms in both accuracy and speed. Compared to MobileFaceNet, RepMLP-FaceRes shows \textbf{4.91}\% higher accuracy and runs 8\% faster (though it has 2.5$\times$ FLOPs), which is obviously a better fit for the high-power devices.

\setlength{\tabcolsep}{4pt}
\begin{table}
	\caption{Results of face recognition on MS1M-V2 and MegaFace. The speed (examples/second) is tested with a batch size of 512 and input 96$\times$96 on the same 1080Ti GPU .}
	\label{table-face}
	\vspace{-0.2in}
	\begin{center}
		\small
		\begin{tabular}{lcccccccc}
			\hline
			Model					&	Acc		&	 Speed		&	FLOPs (M)	&	Params (M) 	\\
			\hline
			MobileFaceNet			&	90.99		&	5002		&	162			&	0.98	\\
			FaceResNet				&	92.97		&	3811		&	1050		&	40.35		\\	
			RepMLP-FaceRes			&	95.90		&	5425		&	406			&	28.32	\\
			\hline
		\end{tabular}
	\end{center}
	\vspace{-0.25in}
\end{table}
\setlength{\tabcolsep}{1.4pt}

\subsection{Semantic Segmentation}

Semantic segmentation is a representative task with translation invariance, as a car may occur at the left or right. We verify the generalization performance of ImageNet-pretrained RepMLP-Res50 on Cityscapes \cite{cityscapes}, which contains 5K finely annotated images and 19 categories. We use the RepMLP-Res50-g4/8 and the original ResNet-50 pretrained with $320\times320$ on ImageNet as the backbones. For the better reproducibility, we simply adopt the official implementation and default configurations \cite{official-pspnet} of PSPNet \cite{pspnet} framework: poly learning rate policy with base of 0.01 and power of 0.9, weight decay of $10^{-4}$ and a global batch size of 16 on 8 GPUs for 200 epochs. Following PSPNet-50, we use dilated conv in c5 of both models and c4 of the original ResNet-50. We do not use dilated conv in c4 of RepMLP-Res50-g4/8 because its receptive field is already large. Since the resolution of c3 and c4 becomes $90\times90$, the Global Perceptron will have 81 partitions of each channel hence more parameters in FC1 and FC2. We address this problem by reducing the output dimensions of the FC1 and the input dimensions of FC2 by 4$\times$ for c3 and 8$\times$ for c4. FC1 are FC2 are initialized randomly, and all the other parameters are inherited from the ImageNet-pretrained model.

Table. \ref{table-seg} shows that the PSPNet with RepMLP-Res50-g4/8 outperforms the Res-50 backbone by 2.21\% in mIoU. Though it has more parameters, the FLOPs is lower and the speed is faster. Of note is that our PSPNet baseline is lower than the reported PSPNet-50 because the latter was customized for semantic segmentation (added two more layers before the max pooling) but ours is not.

\setlength{\tabcolsep}{4pt}
\begin{table}
	\caption{Semantic segmentation on Cityscapes \cite{cityscapes} tested on the \textit{validation} subset. The speed (examples/second) is tested with a batch size of 16 and input 713$\times$713 on the same 1080Ti GPU.}
	\label{table-seg}
	\vspace{-0.2in}
	\begin{center}
		\small
		\begin{tabular}{lccccccc}
			\hline
			Backbone				&	mIoU	&	Speed	&	FLOPs (M)	&	Params (M)	\\
			\hline
			RepMLP-Res50			&	76.58		&	10.43	&	342,696		&	175.41	\\
			ResNet-50				&	74.27		&	10.22	&	350,004		&	46.56		\\
			\hline
		\end{tabular}
	\end{center}
	\vspace{-0.1in}
\end{table}
\setlength{\tabcolsep}{1.4pt}

\section{Conclusion}

An FC has stronger representational capacity than a conv, as the latter can be viewed as a sparse FC with shared parameters. However, an FC has no local prior, which makes it less favored for image recognition. In this paper, we have proposed RepMLP, which utilizes the global capacity and positional perception of FC and incorporates the local prior into FC by re-parameterizing convolutions into it via a simple and platform-agnostic algorithm. From the theoretical side, \textit{viewing conv as a degraded case of FC} opens up a new perspective, which may deepen our understanding of the traditional ConvNets. It should not be left unmentioned that RepMLP is designed for the application scenarios where the major concerns are the inference throughput and accuracy, less concerning the number of parameters.

{\small
\bibliographystyle{ieee_fullname}
\bibliography{repmlpbib}

\begin{thebibliography}{10}\itemsep=-1pt

\bibitem{cao2019gcnet}
Yue Cao, Jiarui Xu, Stephen Lin, Fangyun Wei, and Han Hu.
\newblock Gcnet: Non-local networks meet squeeze-excitation networks and
  beyond.
\newblock In {\em Proceedings of the IEEE/CVF International Conference on
  Computer Vision Workshops}, pages 0--0, 2019.

\bibitem{im2col}
Kumar Chellapilla, Sidd Puri, and Patrice Simard.
\newblock High performance convolutional neural networks for document
  processing.
\newblock In {\em Tenth International Workshop on Frontiers in Handwriting
  Recognition}. Suvisoft, 2006.

\bibitem{chen2018mobilefacenets}
Sheng Chen, Yang Liu, Xiang Gao, and Zhen Han.
\newblock Mobilefacenets: Efficient cnns for accurate real-time face
  verification on mobile devices.
\newblock In {\em Chinese Conference on Biometric Recognition}, pages 428--438.
  Springer, 2018.

\bibitem{cho2017mec}
Minsik Cho and Daniel Brand.
\newblock Mec: memory-efficient convolution for deep neural network.
\newblock In {\em International Conference on Machine Learning}, pages
  815--824. PMLR, 2017.

\bibitem{chollet2017xception}
Fran{\c{c}}ois Chollet.
\newblock Xception: Deep learning with depthwise separable convolutions.
\newblock In {\em Proceedings of the IEEE conference on computer vision and
  pattern recognition}, pages 1251--1258, 2017.

\bibitem{cityscapes}
Marius Cordts, Mohamed Omran, Sebastian Ramos, Timo Rehfeld, Markus Enzweiler,
  Rodrigo Benenson, Uwe Franke, Stefan Roth, and Bernt Schiele.
\newblock The cityscapes dataset for semantic urban scene understanding.
\newblock In {\em 2016 {IEEE} Conference on Computer Vision and Pattern
  Recognition, {CVPR} 2016, Las Vegas, NV, USA, June 27-30, 2016}, pages
  3213--3223. {IEEE} Computer Society, 2016.

\bibitem{cubuk2019autoaugment}
Ekin~D Cubuk, Barret Zoph, Dandelion Mane, Vijay Vasudevan, and Quoc~V Le.
\newblock Autoaugment: Learning augmentation strategies from data.
\newblock In {\em Proceedings of the IEEE conference on computer vision and
  pattern recognition}, pages 113--123, 2019.

\bibitem{deng2009imagenet}
Jia Deng, Wei Dong, Richard Socher, Li-Jia Li, Kai Li, and Li Fei-Fei.
\newblock Imagenet: A large-scale hierarchical image database.
\newblock In {\em Computer Vision and Pattern Recognition, 2009. CVPR 2009.
  IEEE Conference on}, pages 248--255. IEEE, 2009.

\bibitem{ding2019acnet}
Xiaohan Ding, Yuchen Guo, Guiguang Ding, and Jungong Han.
\newblock Acnet: Strengthening the kernel skeletons for powerful cnn via
  asymmetric convolution blocks.
\newblock In {\em Proceedings of the IEEE International Conference on Computer
  Vision}, pages 1911--1920, 2019.

\bibitem{ding2021repvgg}
Xiaohan Ding, Xiangyu Zhang, Ningning Ma, Jungong Han, Guiguang Ding, and Jian
  Sun.
\newblock Repvgg: Making vgg-style convnets great again.
\newblock {\em arXiv preprint arXiv:2101.03697}, 2021.

\bibitem{dosovitskiy2020image}
Alexey Dosovitskiy, Lucas Beyer, Alexander Kolesnikov, Dirk Weissenborn,
  Xiaohua Zhai, Thomas Unterthiner, Mostafa Dehghani, Matthias Minderer, Georg
  Heigold, Sylvain Gelly, et~al.
\newblock An image is worth 16x16 words: Transformers for image recognition at
  scale.
\newblock {\em arXiv preprint arXiv:2010.11929}, 2020.

\bibitem{guo2016ms}
Yandong Guo, Lei Zhang, Yuxiao Hu, Xiaodong He, and Jianfeng Gao.
\newblock Ms-celeb-1m: A dataset and benchmark for large-scale face
  recognition.
\newblock In {\em European conference on computer vision}, pages 87--102.
  Springer, 2016.

\bibitem{han2015learning}
Song Han, Jeff Pool, John Tran, and William Dally.
\newblock Learning both weights and connections for efficient neural network.
\newblock In {\em Advances in Neural Information Processing Systems}, pages
  1135--1143, 2015.

\bibitem{he2016deep}
Kaiming He, Xiangyu Zhang, Shaoqing Ren, and Jian Sun.
\newblock Deep residual learning for image recognition.
\newblock In {\em Proceedings of the IEEE conference on computer vision and
  pattern recognition}, pages 770--778, 2016.

\bibitem{ioffe2015batch}
Sergey Ioffe and Christian Szegedy.
\newblock Batch normalization: Accelerating deep network training by reducing
  internal covariate shift.
\newblock In {\em International Conference on Machine Learning}, pages
  448--456, 2015.

\bibitem{kemelmacher2016megaface}
Ira Kemelmacher-Shlizerman, Steven~M Seitz, Daniel Miller, and Evan Brossard.
\newblock The megaface benchmark: 1 million faces for recognition at scale.
\newblock In {\em Proceedings of the IEEE conference on computer vision and
  pattern recognition}, pages 4873--4882, 2016.

\bibitem{winograd}
Andrew Lavin and Scott Gray.
\newblock Fast algorithms for convolutional neural networks.
\newblock In {\em Proceedings of the IEEE Conference on Computer Vision and
  Pattern Recognition}, pages 4013--4021, 2016.

\bibitem{lian2021mlp}
Dongze Lian, Zehao Yu, Xing Sun, and Shenghua Gao.
\newblock As-mlp: An axial shifted mlp architecture for vision.
\newblock {\em arXiv preprint arXiv:2107.08391}, 2021.

\bibitem{liu2021pay}
Hanxiao Liu, Zihang Dai, David~R So, and Quoc~V Le.
\newblock Pay attention to mlps.
\newblock {\em arXiv preprint arXiv:2105.08050}, 2021.

\bibitem{fft-conv}
Michael Mathieu, Mikael Henaff, and Yann LeCun.
\newblock Fast training of convolutional networks through ffts.
\newblock {\em arXiv preprint arXiv:1312.5851}, 2013.

\bibitem{molchanov2016pruning}
Pavlo Molchanov, Stephen Tyree, Tero Karras, Timo Aila, and Jan Kautz.
\newblock Pruning convolutional neural networks for resource efficient
  inference.
\newblock In {\em 5th International Conference on Learning Representations,
  {ICLR} 2017, Toulon, France, April 24-26, 2017, Conference Track
  Proceedings}. OpenReview.net, 2017.

\bibitem{paszke2019pytorch}
Adam Paszke, Sam Gross, Francisco Massa, Adam Lerer, James Bradbury, Gregory
  Chanan, Trevor Killeen, Zeming Lin, Natalia Gimelshein, Luca Antiga, et~al.
\newblock Pytorch: An imperative style, high-performance deep learning library.
\newblock {\em arXiv preprint arXiv:1912.01703}, 2019.

\bibitem{torch-model}
PyTorch.
\newblock {\em Torchvision Official Models}, 2020.

\bibitem{efficientnet}
Mingxing Tan and Quoc~V Le.
\newblock Efficientnet: Rethinking model scaling for convolutional neural
  networks.
\newblock {\em arXiv preprint arXiv:1905.11946}, 2019.

\bibitem{tolstikhin2021mlp}
Ilya Tolstikhin, Neil Houlsby, Alexander Kolesnikov, Lucas Beyer, Xiaohua Zhai,
  Thomas Unterthiner, Jessica Yung, Daniel Keysers, Jakob Uszkoreit, Mario
  Lucic, et~al.
\newblock Mlp-mixer: An all-mlp architecture for vision.
\newblock {\em arXiv preprint arXiv:2105.01601}, 2021.

\bibitem{touvron2021resmlp}
Hugo Touvron, Piotr Bojanowski, Mathilde Caron, Matthieu Cord, Alaaeldin
  El-Nouby, Edouard Grave, Armand Joulin, Gabriel Synnaeve, Jakob Verbeek, and
  Herv{\'e} J{\'e}gou.
\newblock Resmlp: Feedforward networks for image classification with
  data-efficient training.
\newblock {\em arXiv preprint arXiv:2105.03404}, 2021.

\bibitem{vaswani2017attention}
Ashish Vaswani, Noam Shazeer, Niki Parmar, Jakob Uszkoreit, Llion Jones,
  Aidan~N Gomez, Lukasz Kaiser, and Illia Polosukhin.
\newblock Attention is all you need.
\newblock {\em arXiv preprint arXiv:1706.03762}, 2017.

\bibitem{wang2018additive}
Feng Wang, Jian Cheng, Weiyang Liu, and Haijun Liu.
\newblock Additive margin softmax for face verification.
\newblock {\em IEEE Signal Processing Letters}, 25(7):926--930, 2018.

\bibitem{wang2018non}
Xiaolong Wang, Ross Girshick, Abhinav Gupta, and Kaiming He.
\newblock Non-local neural networks.
\newblock In {\em Proceedings of the IEEE conference on computer vision and
  pattern recognition}, pages 7794--7803, 2018.

\bibitem{xiapatent}
Chunlong Xia.
\newblock Visual task processing method and device and electronic system, Oct.
  2020.

\bibitem{xie2017aggregated}
Saining Xie, Ross Girshick, Piotr Doll{\'a}r, Zhuowen Tu, and Kaiming He.
\newblock Aggregated residual transformations for deep neural networks.
\newblock In {\em Proceedings of the IEEE conference on computer vision and
  pattern recognition}, pages 1492--1500, 2017.

\bibitem{zhang2017mixup}
Hongyi Zhang, Moustapha Cisse, Yann~N Dauphin, and David Lopez-Paz.
\newblock mixup: Beyond empirical risk minimization.
\newblock {\em arXiv preprint arXiv:1710.09412}, 2017.

\bibitem{official-pspnet}
Hengshuang Zhao.
\newblock Official pspnet.
\newblock \url{https://github.com/hszhao/semseg}, 2020.

\bibitem{pspnet}
Hengshuang Zhao, Jianping Shi, Xiaojuan Qi, Xiaogang Wang, and Jiaya Jia.
\newblock Pyramid scene parsing network.
\newblock In {\em 2017 {IEEE} Conference on Computer Vision and Pattern
  Recognition, {CVPR} 2017, Honolulu, HI, USA, July 21-26, 2017}, pages
  6230--6239. {IEEE} Computer Society, 2017.

\end{thebibliography}
}

\clearpage

\section*{Appendix A: RepMLP-ResNet for High Speed}

\begin{figure*}
	\begin{center}
		\includegraphics[width=\linewidth]{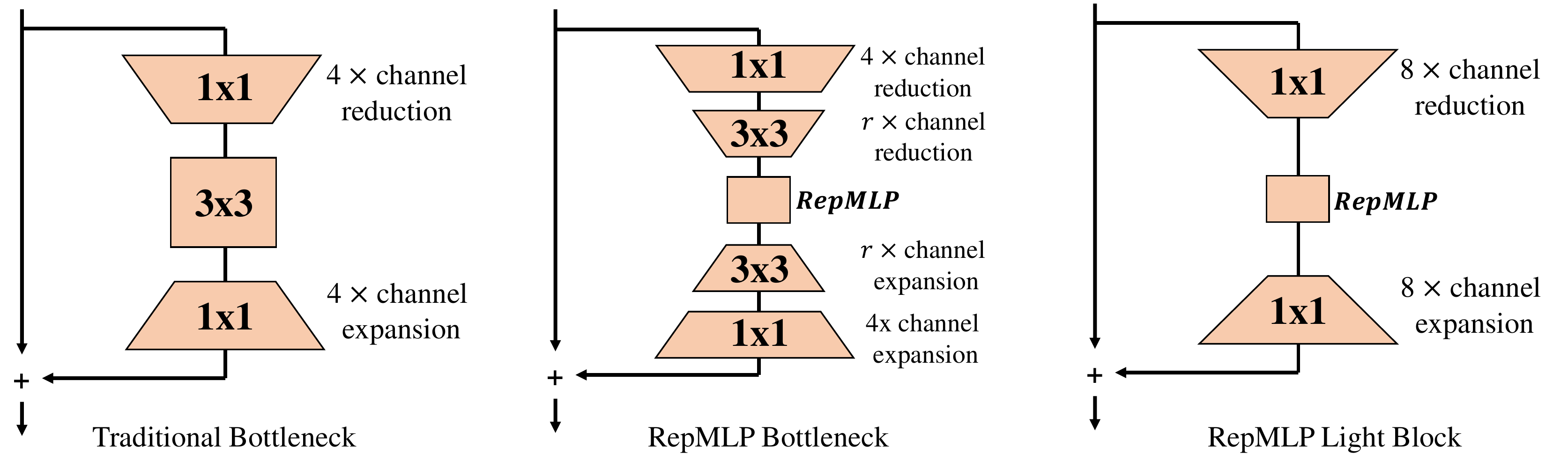}
		\vspace{-0.05in}
		\caption{The original bottleneck, RepMLP Bottleneck and RepMLP Light Block.}
		\label{fig-block}
	\end{center}
	\vspace{-0.2in}
\end{figure*}

The RepMLP Bottleneck presented in the paper is designed to improve the accuracy. Here we present another means of using RepMLP in ResNet for the higher speed. Specifically, we build a \textit{RepMLP Light Block} (Fig. \ref{fig-block}) with no $3\times3$ conv but drastic $8\times$ channel reduction/expansion via $1\times1$ conv before and after RepMLP. Same as the 78.55\%-accuracy RepMLP-ResNet50 reported in the paper, we use $h=w=7$, $g=8$ and three conv branches in the Local Perceptron with $K=1,3,5$. The speed is tested in the same way as all the models reported in the paper. Table. \ref{table-fast} shows that the ResNet with RepMLP Light Block achieves almost the same accuracy as the original ResNet-50 with 30\% lower FLOPs and 55\% faster speed.

Of note is that RepMLP is a building block that can be combined with numerous other structures in various ways. We only present two means for using RepMLP in ResNet, which may not be the optimal. We will make the code and models publicly available to encourage further research.

\setlength{\tabcolsep}{4pt}
\begin{table}
	\caption{ResNet-50 with different blocks in c3 and c4. The speed is in examples/second.}
	\label{table-fast}
	\vspace{-0.25in}
	\begin{center}
		\small
		\begin{tabular}{lccccccc}
			\hline
			Block				&	Top-1 acc	&	Speed	&	FLOPs (M)	&	Params (M)	\\
			\hline
			Original			&		77.19	&	689		&	4089		&	25.53	\\
			RepMLP Bottleneck	&		78.55	&	636		&	3890		&	40.87	\\
			RepMLP Light BLock	&		77.14	&	1074	&	2919		&	57.86	\\
			\hline
		\end{tabular}
	\end{center}
	\vspace{-0.15in}
\end{table}
\setlength{\tabcolsep}{1.4pt}

\section*{Appendix B: Converting Groupwise Conv into FC}

The groupwise case of converting conv into FC is a bit more complicated, which can be derived by first splitting the input into $g$ parallel groups and then converting every group separately. The PyTorch code is shown in Alg. \ref{alg1} and the submitted \textit{repmlp.py} contains an executable example to verify the equivalence. It is easy to verify that with $g=1$ the code exactly implements Eq. 15 in the paper.

\begin{algorithm*}
	\vskip -0.03in
	\caption{PyTorch code for converting groupwsie conv into FC.}
	\label{alg1}
	\begin{algorithmic}
		\STATE {\bfseries Input: C, h, w, g, O, conv\_kernel, conv\_bias}
		\STATE I = torch.eye(C * h * w // g).repeat(1, g).reshape(C * h * w // g, C, h, w)
		\STATE fc\_kernel = F.conv2d(I, conv\_kernel, padding=conv\_kernel.size(2)//2, groups=g)
		\STATE fc\_kernel = fc\_kernel.reshape(O * h * w // g, C * h * w).t()	\# Note the transpose
		\STATE fc\_bias = conv\_bias.repeat\_interleave(h * w)
		\STATE {\bfseries return: fc\_kernel, fc\_bias}
	\end{algorithmic}
	\vskip -0.03in
\end{algorithm*}

\section*{Appendix C: Absorbing BN into FC1}

The BN in Global Perceptron applies a linear scaling and a bias adding to the input. After the matrix multiplication by the FC1 kernel, the added bias is projected and then added onto the bias of FC1. Therefore, the removal of this BN can be offset by scaling the kernel of FC1 and changing the bias of FC1. The code is shown in Alg. \ref{alg2} and the submitted \textit{repmlp.py} contains an executable example to verify the equivalence.

\begin{algorithm*}
	\vskip -0.03in
	\caption{PyTorch code for absorbing BN into FC1.}
	\label{alg2}
	\begin{algorithmic}
		\STATE {\bfseries Input: mean, std, gamma, beta of the BN, fc\_kernel, fc\_bias of FC1}
		\STATE scale = gamma / std
		\STATE avgbias = beta - mean * scale
		\STATE replicate\_times = fc\_kernel.shape(1) // len(avgbias)
		\STATE replicated\_avgbias = avgbias.repeat\_interleave(replicate\_times).view(-1, 1)
		\STATE bias\_diff = fc\_kernel.matmul(replicated\_avgbias).squeeze()
		\STATE fc\_bias\_new = fc\_bias + bias\_diff
		\STATE fc\_kernel\_new = fc\_kernel * scale.repeat\_interleave(replicate\_times).view(1, -1)
		\STATE {\bfseries return: fc\_kernel\_new, fc\_bias\_new}
		
	\end{algorithmic}
	\vskip -0.03in
\end{algorithm*} 

\end{document}